\documentclass[lettersize,journal]{IEEEtran}

\usepackage{hyperref}       
\usepackage[caption=false,font=normalsize,labelfont=sf,textfont=sf]{subfig}

\hypersetup{
    colorlinks = true,
    citecolor = blue,
    linkcolor = blue,
    urlcolor = blue,
    bookmarksopen = true,
    filecolor=blue
}

\usepackage{cite}
\usepackage{algorithmic}
\usepackage{algorithm}
\usepackage{array}
\usepackage{url}            
\usepackage{booktabs}       
\usepackage{verbatim}
\usepackage{stfloats}
\usepackage{graphicx}
\usepackage{amsmath,amsfonts,amssymb}
\usepackage{nicefrac}       
\usepackage{xcolor}         
\usepackage{multirow}
\usepackage{dsfont}

\DeclareMathOperator{\MSE}{MSE}

\DeclareMathOperator{\MaskedL2}{MaskedL2}
\DeclareMathOperator{\Mixed}{Mix}
\DeclareMathOperator{\Physical}{Physical}

\providecommand{\birow}[1]{\multirow{2}{*}{#1}}
\providecommand{\mc}{\multicolumn}
\providecommand{\bosym}{\boldsymbol}

\begin{document}

\title{PowerFlowNet: Power Flow Approximation Using Message Passing Graph Neural Networks}

\author{Nan Lin~\IEEEmembership{Student Member,~IEEE}, Stavros Orfanoudakis~\IEEEmembership{Student Member,~IEEE}, Nathan Ordonez Cardenas, Juan~S.~Giraldo,~\IEEEmembership{Member,~IEEE}, Pedro~P.~Vergara,~\IEEEmembership{Senior Member,~IEEE}
\thanks{
This work used the Dutch national e-infrastructure with the support of the
SURF Cooperative using grant no. EINF-6262. This publication is part of the project ALIGN4energy (with project number NWA.1389.20.251) of the research programme NWA ORC 2020 which is (partly) financed by the Dutch Research Council (NWO). Stavros is by the HORIZON Europe Drive2X Project 101056934.}
\thanks{
Nan Lin, Stavros Orfanoudakis, and Pedro P. Vergara are with the Intelligent Electrical Power Grids (IEPG) Section, Delft University of Technology, Delft, The Netherlands (emails: {n.lin, s.orfanoudakis, p.p.vergarabarrios}@tudelft.nl). 
}
\thanks{Nathan Ordonez Cardenas is with the Computer Science Department, Delft University of Technology, Delft, The Netherlands (email: {n.a.ordonezcardenas}@student.tudelft.nl). }%
\thanks{Juan S. Giraldo is with the Energy Transition Studies Group, Netherlands Organization for Applied Scientific Research (TNO), Amsterdam, The Netherlands (email: juan.giraldo@tno.nl)}
\thanks{The first two authors contributed equally.}
}%

\markboth{IEEE TRANSACTIONS ON Smart Grid~-,~Vol.~-, No.~-, ~-}%
{Lin \MakeLowercase{\textit{et al.}}: PowerFlowNet: Power Flow Approximation Using Message Passing Graph Neural Networks}


\maketitle

\begin{abstract}
Accurate and efficient power flow (PF) analysis is crucial in modern electrical networks' operation and planning. Therefore, there is a need for scalable algorithms that can provide accurate and fast solutions for both small and large scale power networks. As the power network can be interpreted as a graph, Graph Neural Networks (GNNs) have emerged as a promising approach for improving the accuracy and speed of PF approximations by exploiting information sharing via the underlying graph structure. In this study, we introduce PowerFlowNet, a novel GNN architecture for PF approximation that showcases similar performance with the traditional Newton-Raphson method but achieves it 4 times faster in the IEEE 14-bus system and 145 times faster in the realistic case of the French high voltage network (6470rte). Meanwhile, it significantly outperforms other traditional approximation methods, such as the DC power flow, in terms of performance and execution time; therefore, making PowerFlowNet a highly promising solution for real-world PF analysis. Furthermore, we verify the efficacy of our approach by conducting an in-depth experimental evaluation, thoroughly examining the performance, scalability, interpretability, and architectural dependability of PowerFlowNet. The evaluation provides insights into the behavior and potential applications of GNNs in power system analysis.  
\end{abstract}


\section{Introduction}
\IEEEPARstart{T}{he} complexity of electrical power systems is continuously rising, largely attributed to the substantial integration of decentralized renewable energy resources. Within this context, power flow (PF) stands as a fundamental challenge in ensuring the stability of power systems, playing a pivotal role in both the operational management and long-term planning of electrical networks. At its core, PF is a mathematical problem that revolves around determining the voltages at various buses, a task accomplished by solving a set of nonlinear equations, which are inherently linked to the network configuration, load distribution, and generation characteristics~\cite{Albadi19}. Traditional methods, such as Newton-Raphson~\cite{Costa1999DevelopmentsIT}, the Gauss-Seidel~\cite{Eltamaly2017LoadFA}, and the fast-decoupled methods, have excellent accuracy and convergence properties but scale slowly with larger power systems, particularly in long-term planning of national grids with thousands of buses~\cite{7725984}. In contrast, the DC power flow (DCPF) technique~\cite{seifi2011appendix} simplifies this problem into a linear one by making assumptions about the voltage magnitude, but at the cost of accuracy. Consequently, there is a need for innovative algorithms capable of efficiently solving the PF for significantly larger networks. These novel algorithms should aim to balance the need for speed without compromising accuracy, ultimately facilitating the streamlined operation and planning of large real-world electrical grids.

Modernizing the grids with the addition of accurate smart metering and data acquisition systems has enabled the development of Machine Learning (ML) methods for accurate and efficient power system analysis~\cite{Liu2019Data-drivenApproach2}. 
ML PF methods use historical operation data gathered from the metering infrastructure and try to approximate the power system state based on them.
In~\cite{Guo2022Data-DrivenApproach2}, the authors effectively transformed the nonlinear PF relationship into a linear mapping within a higher-dimensional state space, resulting in a substantial improvement in the precision of the calculation process. Similarly, Chen et al.~\cite{9662079} mitigate the errors of model-based PF linearization approaches by approximating the nonlinear PF equations in a data-driven manner. In~\cite{Chen2022Data-drivenEquations}, the data-driven PF method comprises two distinct stages: offline learning and online computing. During the offline learning stage, a learning model is developed utilizing the proposed exact linear regression equations, which is subsequently solved by applying the ridge regression method to mitigate the impact of data collinearity. Subsequently, the need for nonlinear iterative calculations is obviated in the online computing stage, thereby streamlining the computation process. Furthermore, Liu et al.~\cite{8924608}  
reformulated PF into a regression model, leveraging the structural attributes of AC power flow equations, including Jacobian matrix-guided constraints, to substantially reduce the search space. Overall, all these simple ML methods have outperformed the DCPF method~\cite{Ramasamy2022PowerConditions}; however, the scalability issues still do not allow for the application of these methods in large power systems.

Numerous deep learning techniques have been developed, promising to find more scalable solutions. Many studies emphasize the use of physics-informed methodologies that leverage the inherent characteristics of the problem. For example, Hu et al.~\cite{9216092} incorporate an auxiliary task for PF model reconstruction, wherein the neural network (NN) based PF solver is effectively regularized by encoding varying levels of Kirchhoff's laws and system topology into the reconstructed PF model, consequently ensuring adherence to physical laws and constraints. In~\cite{Yang2020FastApproach2}, the training process of the NN is enhanced through the physical PF equations, such as incorporating branch flows as a penalty term in the NN's objective function and simplifying the backpropagation update gradients based on the transmission grid's physical characteristics. Similarly, Li et al.~\cite{10138375} use NNs to estimate the distribution system state equations by creating virtual nodes to represent buses without any smart metering. An alternative approach is to enhance a classical solver (Gauss-Newton) by training an NN to serve as a regularizer~\cite{Yang2020PowerSS}. 
However, it is important to note that classic deep learning approaches do not exploit information sharing via the underlying graph structure of power system, leading to limited generalization capabilities and efficiency in real-world large-size network scenarios. 

GNNs~\cite{DBLP:conf/iclr/KipfW17} have garnered significant attention in recent years due to their capacity to leverage graph-structured data by aggregating information from neighboring nodes. In practice, GNNs are efficient because the same parameters can be used for every node. 
Consequently, researchers have explored diverse GNN architectures to tackle the PF problem because the topology of a power system can be naturally interpreted as a graph. In particular, GNNs have been successful in various tasks related to power systems~\cite{GNNSurvey}, such as fault detection, time-series prediction, power flow calculation, and data generation.
Wu et al.~\cite{Wu2023Spatio-TemporalAlgorithms2} used the power flow equations to derive a novel representation of the power system topology, which can be utilized in GNNs to improve performance in power grid state forecasting and voltage control through reinforcement learning. In~\cite{10167837}, a weakly supervised learning approach is employed based on the PF equations that do not require labeled data, however, at the cost of accuracy compared to fully supervised methods.
Furthermore, Pagnier et al.~\cite{pagnier2021PhysicsInformedGraphicalNeural} achieved improvements on the previous method by implementing a specific encoder-decoder model loss\footnote{In this paper, we distinguish the model loss for training a neural network and the line loss of the power system.} for the prediction of effective PF model parameters. 
Meanwhile, \cite{donon2020NeuralNetworksPower} improves upon an iterative-based GNN approach 
by employing a purely physics-constrained model loss. 
%
Despite the proliferation of various studies in this domain, little attention has been directed toward investigating the effectiveness of proposed methodologies for large power networks. Consequently, a comprehensive evaluation and comparison of traditional and GNN-based approaches concerning accuracy, computational complexity, scalability, and interpretability are imperative to discern the most promising techniques for PF analysis.

To address the scalability and other aforementioned real-world application concerns, we propose PowerFlowNet. PowerFlowNet is a novel GNN architecture specifically designed to leverage electrical power networks' structural characteristics and interconnectedness, enabling efficient approximation of the PF. In practice, PowerFlowNet transforms the PF into a GNN node-regression problem by representing each bus as a node and each transmission line as an edge while maintaining the network's connectivity. The key advantage of PowerFlowNet lies in its significantly lower execution time, regardless of the network size; thus, enabling it to be applied for realistic power grid planning and operation processes. Specifically, our thorough experimental evaluation demonstrates PowerFlowNet's accuracy and execution speed compared to traditional methods, such as DCPF, while its performance is similar to the most trusted Newton-Raphson method but with significantly lower execution time. During planning and operation with distributed energy resources, thousands of PF calculations are needed because of the number of possible scenarios. In such cases, accuracy and low execution time are indispensable. 
Additionally, an in-depth analysis of the proposed approach's scalability, architectural dependability, and interpretability is conducted, aiming to provide valuable insights into the broader applicability of GNN algorithms in the power system domain. 

In detail, the key contribution of this paper is summarized as follows:
\begin{itemize}
     \item We introduce a novel GNN architecture for PF approximation with significantly lower executing time (during operation) and comparable results compared to the Newton-Raphson method. Its distinctiveness, compared to existing PF GNN approaches, lies in its adept utilization of the capabilities from message passing GNNs~\cite{gilmer2017NeuralMessagePassinga} and high-order GCN~\cite{du2017topology}, in a unique arrangement called PowerFlowConv, for handling a trainable masked embedding of the network graph. This innovative approach renders the proposed GNN architecture scalable, presenting an effective solution for the PF problem for larger size networks.
    
\end{itemize}


\section{GNNs and Classic PF Algorithms} \label{chap2}
This section provides an overview of the fundamental background of GNNs and the methodologies they employ. Furthermore, it delves into the traditional techniques used in PF analysis, including the DCPF method. 

\subsection{Graph Neural Networks}
GNNs are a class of ML models designed to operate on graph-structured data, which can represent complex relationships and interactions between entities~\cite{Kipf2016Semi-SupervisedNetworks}. GNNs have gained significant attention due to their ability to capture and leverage the structural information inherent in graphs. In detail, a graph $\mathcal{G}(\mathcal{N},\mathcal{E})$ can be defined as a set of $\mathcal{N}$ nodes and $\mathcal{E}$ edges with node features\footnote{For disambiguitation, we always represent the node and edge features with letter $x$ and the line impedance with letter $z$.} $\bosym x_i \in{\mathbb{R}^{F}}$ and edge features $\bosym x^e_{i,j}\in{\mathbb{R}^{F_e}}$, where $i \in \mathcal{N}$ and the pair $(i,j) \in \mathcal{E}$. Additionally, GNNs can use regular, fully connected layers to help them understand and work with the structure of these graphs.

\subsubsection{Message Passing NNs}
\begin{figure}[!t]
    \centering
    \includegraphics[width=0.9\linewidth]{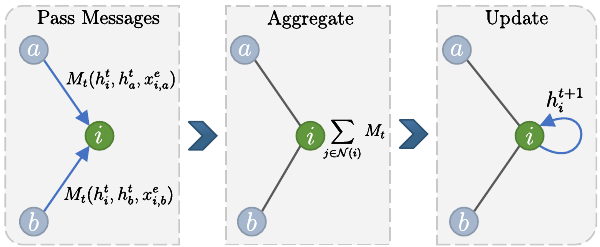}
    \caption{A message passing step of node $i$ consisting of message creation, aggregation, and update of the hidden state.}
    \label{fig:message-passing}
\end{figure}

At the heart of GNNs lies the concept of message passing \cite{gilmer2017NeuralMessagePassinga}. Message passing allows information to propagate through the graph by iteratively aggregating and updating node features based on their neighborhood relationships. In each message passing step, each node receives information from its neighbors, performs local computations, and then transmits updated information to its neighbors (Fig.~\ref{fig:message-passing}).
Specifically, the message passing phase can run iteratively for $T$ steps. The nodes features in the intermediate layers are often called hidden states. For every node $i$, its hidden state $\bosym{h}^t_i$ is updated based on the message function $M_t( \cdot)$ and the node update function $U_t( \cdot)$. Therefore, for node $i$ with neighbors $\mathcal{N}(i)$, each iterative message update $\bosym{v}_i^{t+1}$ is
\begin{equation}
    \bosym{v}_i^{t+1} = \sum_{j \in \mathcal{N}(i)} M_t( \bosym h_i^t, \bosym h_j^t, \bosym x^e_{i,j}),
    \label{eq:mp1}
\end{equation}
\begin{equation}
    \bosym h_i^{t+1} = U_t( \bosym h_i^t, \bosym v_i^{t+1}).
\end{equation}
As shown in \eqref{eq:mp1}, the edge features $\bosym x^e_{i,j}$ can also be included in the message creation step.
After the message passing phase ends, the final feature vector for the whole graph is denoted as the output of a function $R(\cdot)$
\begin{equation}
    \hat{\bosym y}(\mathcal{G}) = R(\{\bosym h_i^{\mathsf{T}} | i \in \mathcal{G} \}).
\end{equation}
Notice that the functions $M_t(\cdot), U_t(\cdot)$, and $R( \cdot)$ are designed by the user and should be fully differentiable so that they can be trained and used by graph ML tasks.
Overall, message passing enables all nodes to gather and integrate information from their local context, incorporating both the features of neighboring nodes and the graph structure.

\subsubsection{Graph Convolutional Networks}
One other popular type of GNN is the GCNs~\cite{DBLP:conf/iclr/KipfW17}, which introduces convolution operations on graphs. Similar to convolution neural networks (CNNs) for regular grid-like data, GCNs apply filters to node features and aggregate information from neighboring nodes. By leveraging the local connectivity of the graph, GCNs can capture both node-level patterns and the information of the $K$-hop neighbors. In particular, a graph convolutional layer of order $K$ is defined as
\begin{equation}
    \hat{\bosym{y}}(S,X) = \sigma \left( \sum^{K - 1}_{k=0} S^k X W \right).
\end{equation}
Here, a GCN layer is a function of the graph shift operator $S\in \mathbb{R}^{N\times N}$, node features $X \in \mathbb{R}^{N\times F}$, and trainable weight matrix $W\in{\mathbb{R}^{F\times F}} $. Matrix $S$ can be the adjacency matrix $A$, or a more complex representation such as the Laplacian transformation $L$. Moreover, $\sigma$ represents a non-linear function, such as rectified linear units (ReLU), which is employed to enable the approximation of every target non-linear function.
The hyper-parameter $K$ holds significance as it dictates the $K$-hop\footnote{Nodes with distance $K$ from the original node.} neighboring nodes are taken into account during each convolution. Consequently, a larger value of $K$ entails the inclusion of more nodes.

\subsubsection{Topology Adaptive GCN}
\label{sec:tagconv}
Many variants of the GCN layer have been proposed with one of them being the topology adaptive graph convolution operator (TAGConv)~\cite{du2017topology}. TAGConv addresses a limitation of traditional GCN by adaptively learning the importance of different neighboring nodes during message passing. TAGConv assigns weights to the neighbors based on their relevance to the target node. Specifically, a TAGConv layer of order $K$ is defined as
\begin{equation}
    \hat{\bosym y}(S,X) = \sigma \left( \sum^{K - 1}_{k=0} S^k X W_k \right).
    \label{eq:tagconv}
\end{equation}
Here, the trainable weight matrix $W_k\in{\mathbb{R}^{F\times F}}$ is different for every $k$.
This adaptivity enables TAGConv to effectively capture the local structure and important dependencies in the graph.

\subsection{Classic Power Flow Analysis Methods}
\label{sec:pf}
The PF problem is mathematically formulated as solving a set of non-linear equations. A power system has a set of buses (nodes) $\mathcal{N}$ and a set of transmission lines (undirected edges) $\mathcal{E} \subseteq \mathcal{N}\times \mathcal{N}$. For each node $i \in \mathcal{N}$, the PF equations are derived from Kirchhoff's current law and are expressed in complex form as
\begin{equation}
\label{eq:power-flow-equations}
    P^{\text{L}}_i - P^{\text{G}}+ \text{j}(Q^{\text{L}}_i - Q^{\text{G}})
        = \sum_{j\in\mathcal{N}(i)}{
            \Dot{V}_i\left(\frac{\Dot{V}_j-\Dot{V}_i}{z_{i,j}}\right)^*
        }
\end{equation}
where $\text{j}$ is the imaginary unit, italic $i, j$ refer to bus $i$ and $j$, $P_i$ and $Q_i$ are the active and reactive power drawn from bus $i$ to the ground, $\Dot{V}_i=V^{\text{m}}_i\angle \theta_i$ and $\Dot{V}_j=V^{\text{m}}_j\angle \theta_j$ are the voltage phasors at bus $i$ and $j$, $z_{i,j}$ is the impedance of line $(i,j)\in \mathcal{E}$. At each node $i$, two of $P_i$, $Q_i$, $V^{\text{m}}_i$, $\theta_i$ are unknown. For a system with $\left|\mathcal{N}\right|=N$ nodes, there are $2N$ real-valued equations and $2N$ unknown real-valued variables. 

\begin{figure*}[t]
     \hfill
     \subfloat[Standard IEEE 9-case. \label{fig:pf_a}]{
         \centering
         \includegraphics[width=0.33\textwidth]{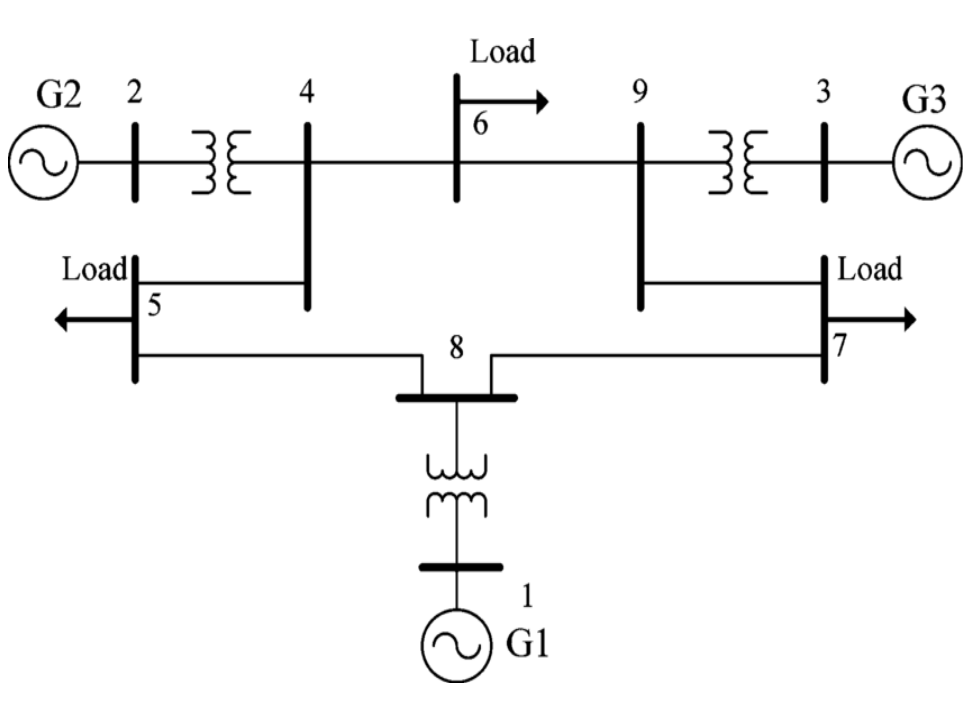}}
     \hfill
     \subfloat[GNN equivalent of the IEEE 9-case. \label{fig:pf_b}]{
         \centering
         \includegraphics[width=0.45\textwidth]{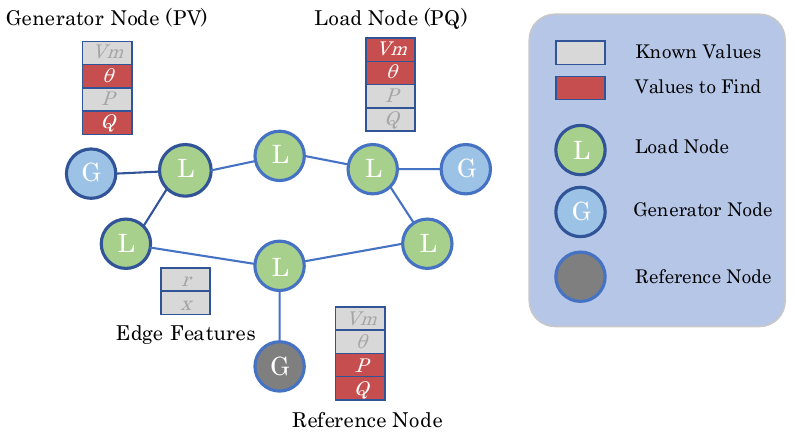}}
         \hfill
        \caption{Interpreting the PF problem of the IEEE 9-case into a GNN node regression problem. In detail, Fig.~\ref{fig:pf_a} shows the standard IEEE 9-case, and Fig.~\ref{fig:pf_b} illustrates how each bus is transformed into a load, generator, or reference node while keeping the same line connectivity. Each node has different known and unknown features $\bosym x = (V^m, \theta, P, Q)$ depending on its type. The ultimate goal of this GNN problem is to approximate all the features for all nodes, thus solving the PF problem.}
        \label{fig:problem-formulation}
\end{figure*}

\subsubsection{DC Power Flow Approximation}
DCPF is an approximation of the ACPF. 
In DCPF, the voltage is assumed constant at all buses; therefore, the voltage and reactive power differences are $0$, so the $\Delta V^m$ and $\Delta Q$ terms are neglected~\cite{seifi2011appendix}. Subsequently, the PF equations can be further simplified to a linear problem that does not require an iterative solution.
In detail, the relation between the active power ($P_i$) and the voltage angle ($\theta_i$) of each bus is
\begin{equation}
    P_i
        = \Re\left[\sum_{j\in\mathcal{N}(i)}{
            \Dot{V}_i\left(\frac{\Dot{V}_j-\Dot{V}_i}{z_{i,j}}\right)^*
        }\right] \approx \sum_{j\in\mathcal{N}(i)}{
            \frac{\theta_j-\theta_i}{\Im{[z_{i,j}]}}
        }. 
        \label{eq:dc}
\end{equation}

\section{PowerFlowNet Framework}\label{chap3}
This section presents a thorough explanation of PowerFlowNet, accompanied by our interpretation of PF formulation in the domain of GNN algorithms.

\subsection{Interpreting Power Flow as a GNN Problem}
The primary objective of a PF study is to determine the voltage magnitudes and angles for a given load, generation, and network state, facilitating subsequent calculations of line flows and power losses. This problem lends itself seamlessly to a GNN approach, as the electrical network topology can be represented as an undirected graph $\mathcal{G}(\mathcal{N},\mathcal{E})$, where the set of nodes $\mathcal{N}$ corresponds to all the buses and the set of edges $\mathcal{E}$ represents power lines. The node features, denoted as $X\in \mathbb{R}^{N\times F}$, encompass various attributes for each bus, including voltage magnitude ($V_i^{\text{m}}$), voltage angle ($\theta_i$), active power ($P_i$), and reactive power ($Q_i$). Similarly, the edge features, denoted as $X^e \in \mathbb{R}^{E\times F_e}$, encompass the resistance ($\Re[z_{i,j}]$) and reactance ($\Im[z_{i,j}]$) attributes for each line $(i,j) \in \mathcal{E}$, where $\Re[\cdot]$ and $\Im[\cdot]$ are the real and imaginary part of a complex variable, respectively.

Illustrated in Fig.~\ref{fig:problem-formulation}, the transformation of a transmission network into a graph representation involves mapping the buses to nodes while preserving the interconnectedness of the lines. The nodes are categorized into three types: a) load (PQ) nodes where the $P_i$ and $Q_i$ are known, b) generator (PV) nodes where $V^{\text{m}}_i$ and $P_i$ are known, and c) a single reference node where both $V^{\text{m}}_i$ and $\theta_i$ are known. The edge features $X^e$ are assumed to be given. The objective of this task is to predict the missing node features for every node $i \in \mathcal{N}$, e.g., for a PQ node, we need to predict the $V^m$ and $\theta_m$, thereby rendering this problem as a node regression task.

\subsection{PowerFlowNet Architecture}
\begin{figure*}[!t]
    \centering
    \includegraphics[width=\textwidth]{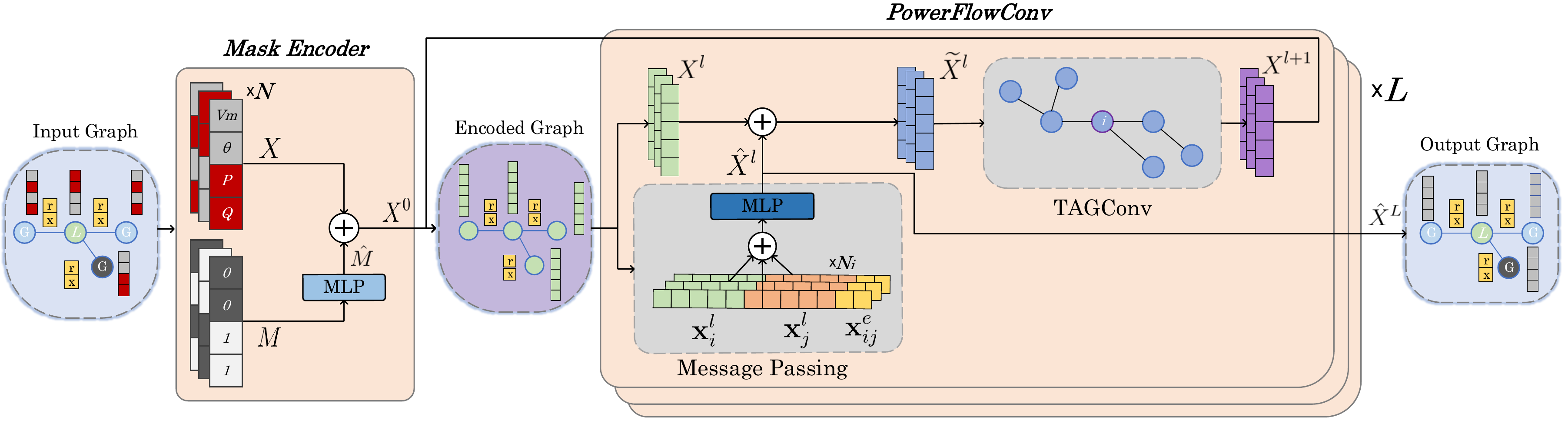}
    \caption{The PowerFlowNet model architecture consists of a mask encoder and $L$ PowerFlowConv layers. The input graph with incomplete feature information is fed node-by-node to the mask encoder to generate encoded graph features, where each node $n \in \mathcal{N}$ with $(\bosym x_i, \bosym m_i)$. Then, the encoded graph features are processed by a series of $L$ sequential PowerFlowConv layers, each comprising a 1-step message passing and a high-order TAGConv. Finally, the complete output graph is produced.}
    \label{fig:model-arch}
\end{figure*}

PowerFlowNet is a GNN approach that reconstructs every node's full feature vector $\hat{\bosym x}_i =$ $(V_i^{\text{m}}, \theta_i, P_i, Q_i)$ given partial information of the problem, such as the adjacency matrix of the graph $A$, the known node features ${X} = [{\bosym x}_1, \ldots,{\bosym x}_N]^\mathsf{T}$, where the unknown features are filled with $0$, and the edge features ${X^e} = [{\bosym x^e}_{i,j}]^\mathsf{T}$ for each line $(i,j) \in \mathcal{E}$. Our proposed model consists of a mask encoder and a stack of our novel Power Flow Convolutional layers (PowerFlowConv). Initially, the mask encoding layers shift the input features to distinguish known and unknown features using a feature mask. Then, as illustrated in Fig.~\ref{fig:model-arch}, the encoded graph features are fed to the stack of Power Flow Convolutional operations consisting of a unique arrangement of message passing and TAGConv layers. This way, information from every node and edge is aggregated so that complete feature matrix $\hat{X} = [\hat{\bosym x}_1, \ldots,\hat{\bosym x}_N]^\mathsf{T}$ is predicted. A detailed explanation of PowerFlowNet's components and procedures is presented next.

\subsubsection{Mask Encoder} \label{sec:me}
In the PF problem, each node has different known and unknown features. The goal is to predict the unknown features while keeping the known ones unchanged. This means that our NN should know which features must be predicted. Consequently, for every input node with feature vector $\bosym x_i$ we create a binary mask $\bosym m_i \in \mathbb{R}^F $ where $0$ represents the known and $1$ the unknown features. For example, the mask of a load (PQ) node with a feature vector $\bosym x_i = (V^m_i,\theta_i, P_i, Q_i)$ and unknown values $V^m$ and $\theta$, will be $\bosym m_i = (1,1,0,0)$. Additionally, 
we propose using a mask encoder that learns to represent different types of nodes (See Mask Encoder block in Fig.~\ref{fig:model-arch}). In practice, it consists of two fully connected layers that map a binary mask to a continuous-valued vector. Notably, a fixed (not learned) mask embedding can also be used, but our mask encoder can learn a more flexible mask representation that improves the final performance. The exact mathematical operation for every $m_i \in M$, where $M \in \mathbb{R}^{N\times F}$, is 
\begin{equation}
    \hat{ \bosym m}_i =  W_1\sigma(W_0 \bosym m_i + b_0) + b_1, \; \forall i\in \mathcal{N},
\end{equation}
which is a function $\bosym{m}_i \in \{0,1\}^{F} \rightarrow \bosym{\hat{m}}_i \in\mathbb{R}^{F}$, and the weight matrices $W_0, W_1$ and the biases $b_0, b_1$ are the trainable parameters. Finally, to produce the encoded graph features $X^l$, we shift the input node features $X$ with the learned representation $\bosym{\hat{m}}_i$ as
\begin{equation}  
    \bosym x_i^{0} = \bosym x_i + \hat{ \bosym m}_i , \; \forall i\in \mathcal{N}.
\end{equation}

\subsubsection{Power Flow Convolutional Layer} The second part of our proposed architecture consists of $L$ connected PowerFlowConv layers that sequentially process the encoded graph features and predict the final feature matrix $\hat{X}^L$, where $\hat{X}^L = [\hat{\bosym x}^L_1, \ldots,\hat{\bosym x}^L_N]^\mathsf{T}$ and $\hat{\bosym x}_i^L =$ $(V_i^{\text{m}}, \theta_i, P_i, Q_i)$, $\forall i \in \mathcal{N}$. Each PowerFlowConv layer consists of an initial one-hop message passing step and a $K$-hop TAGConv (See~\ref{sec:tagconv}), which learns to automatically extract information from the edge features ${\bosym x}_{ij}^e$, i.e., the line resistance and reactance, and incorporates it in the neighboring node features $\bosym{x}_i^l$.
Message passing exploits information from the edge features, and TAGConv aggregates node features in a large neighborhood.
However, in the PF formulation, both line characteristics and node states play an important role.
Therefore, by a combination of both techniques 
the proposed architecture can provide high-quality PF approximations.

More specifically, we calculate the message passed to node $\forall i\in \mathcal{N}$ as 
\begin{equation}
    \hat{\bosym x}^l_i = W^l_{\textit{MP}_1} \cdot \sigma \left( W^l_{\textit{MP}_0}  \sum_{j \in \mathcal{N}(i)} \langle \bosym x_i^{l}, \bosym x_j^{l}, \bosym x_{i,j}^{e} \rangle + b^l_{\textit{MP}_0}\right) + b^l_{\textit{MP}_1},
\end{equation}
where $\langle x_i^{l},x_j^{l},x_{i,j}^{e} \rangle$ is a concatenated vector of the described vectors fed to a two-layer Multilayer Perceptron (MLP) with an in-between ReLU activation function. Here again, the weight matrices and bias vectors $W^l_{\textit{MP}_1},b^l_{\textit{MP}_1},W^l_{\textit{MP}_0},b^l_{\textit{MP}_0}$ are the trainable parameters of the MLP of layer $l$.

Afterward, we update the graph features $X^l$ by summing it up with the message $\hat{X}^l$, i.e. $\hat{X}^l \leftarrow X^l + \hat{X}^l$. Then, the signals $\hat{X}^l$ are processed by the TAGConv layer to generate the new encoded graph features $X^{l+1}$ as in
\begin{equation}
    X^{l+1} = \sigma \left( \sum_{k=0}^{K-1}        (D^{-\frac{1}{2}}AD^{-\frac{1}{2}})^k \hat{X}^l W^l_k \right).
\end{equation}
In $(D^{-1/2}AD^{-1/2})$, the adjacency matrix $A$ of the graph is normalized by the diagonal degree matrix $D$. As shown in~\cite{du2017topology}, the normalized adjacency matrix is often used in GNNs to provide computational stability guarantees. Then, the encoded graph features generated by the $l$\textit{-th} PowerFlowConv layer are passed to the next layer until $l = L$.
Finally, the message passing step is used in the last layer while the TAGConv step is discarded, as depicted in Fig.~\ref{fig:model-arch}.


\subsubsection{Model Loss Functions}

The selection of the appropriate model loss function is important in generating high-quality PF approximations. Usually, purely physics-based model losses are used in existing ML approaches for PF~\cite{10167837,donon2020NeuralNetworksPower}. However, using only physical model losses is insufficient due to the non-linearities of the PF problem. We propose to utilize the Mean Squared Error (MSE) for the training of PowerFlowNet and develop the Masked L2 Loss as a better evaluation metric. In detail, MSE is defined as
\begin{equation}
    \MSE{(\bosym y_i, \hat{\bosym x}_i)} = || \bosym y_i - \hat{\bosym x}_i ||^2_2
    \label{eq:mse_loss}
\end{equation}
where $\bosym y_i$ represents the real values of the features and $\hat{\bosym x}_i$ are the predicted values of node $i$.

To get a more precise value of the actual error in predicting only the unknown features, we developed the Masked L2 loss. This loss function is similar to MSE but only calculates the error for the unknown features, 
\begin{equation}
    \MaskedL2{(\bosym y_i, \hat{\bosym x}_i)} = || \bosym{m}_i \circ (\bosym y_i - \hat{\bosym x}_i) ||^2_2.
\end{equation}
Here, the $\circ$ is the element-wise multiplication operator, and $\bosym{m}_i$ is the binary mask of node $i$ indexing only the features of interest, as defined in Section~\ref{sec:me}. 

\section{Experimental Evaluation}\label{chap5}

This section presents the experimental setup employed in this study, followed by a comprehensive evaluation aimed at elucidating the strengths and weaknesses of the PowerFlowNet model. To gain deeper insights into the efficacy of PowerFlowNet\footnote{PowerFlowNet code, datasets, and trained models can be found at \url{https://github.com/StavrosOrf/PoweFlowNet} and \url{https://github.com/distributionnetworksTUDelft/PoweFlowNet}.} as a GNN approach for PF approximation, the conducted ablation studies focus on four distinct aspects of the proposed approach: performance, interpretability, scalability, and architectural dependability.

\subsection{Experimental Setup}
The evaluation of PowerFlowNet's performance requires the consideration of the varied attributes present in each power network, encompassing differences in size and topology. To address this requirement, three distinct power network cases are evaluated, namely IEEE case-14, IEEE case-118, and case 6470rte~\cite{Josz2016ACPF}, based on their contrasting characteristics. The case with 14 nodes represents a minimal grid topology characterized by a limited number of connections, while the 118 case exhibits a more intricate configuration. Moreover, the 6470rte case accurately emulates the scale and intricacy of the French very high voltage and high voltage transmission network. 
We evaluate PowerFlowNet in this realistic large-scale network to fully show its scalability potential. Notice, however, that the proposed GNN model is topology dependent; hence, once trained, it cannot be used in unseen cases of topology changes, e.g., N-1 contingency, and topology reconfiguration, without any additional training or data samples. Nevertheless, an assessment of its capabilities to generalize to different networks is presented. 

\subsubsection{Dataset Generation}
All aforementioned cases were modeled in PandaPower~\cite{pandapower.2018} and solved using the Newton-Raphson method. For each case, a total of over $30.000$ distinct scenarios were generated by perturbing the default case files. Following the same approach as~\cite{donon2020NeuralNetworksPower}, the generation of each sample involved sampling from random uniform or normal distributions. Specifically, the resistance ($\Re[z_{i,j}]$ in $\Omega$) and reactance ($\Im[z_{i,j}]$ in $\Omega$) values of each line were uniformly selected within $80\%$ and $120\%$ of their original values. In the case of generators, the voltage magnitude ($V^m$) was uniformly set within the range of $[1.00, 1.05]$ per unit, while the initial active powers ($P_g$ in $\text{MW}$) were sampled from a normal distribution $\mathcal{N}(P_g, 0.1|P_g|)$, where the mean corresponds to the original active power and the standard deviation is $10\%$ of it. Lastly, for buses with loads, their active power ($P$ in $\text{MW}$) and reactive power ($Q$ in $\text{MVAR}$) were randomly sampled from normal distributions $\mathcal{N}(P, 0.1|P|)$ and $\mathcal{N}(Q, 0.1|Q|)$, respectively. These samples were then partitioned, allocating $50\%$ for training, $20\%$ for validation, and $30\%$ for testing purposes. 

\subsubsection{Training \& Evaluation}
To train the models, the AdamW optimizer~\cite{Loshchilov2017DecoupledWD} was employed, using a learning rate of $0.001$.
The training process encompassed up to $2000$ epochs (for large models) and $1000$ epochs (for medium and small networks), with each epoch comprising a batch size of 128 samples. In our standard implementation, we use 4 PowerFlowConv layers ($L=4$) of order 3 ($K=3$) to ensure that the computation subgraph of every node covers the majority of the input graph. Additionally, we use $128$ nodes for the hidden dimension and a dropout rate of $0.2$ at the end of each layer to reduce overfitting.
The training time is not considered in the measurement of the execution time since the ML model needs to be trained only once.
We train the models with MSE loss as defined in Equation~\ref{eq:mse_loss}. For assessment purposes, we also trained with the physical loss and the mixed loss ($w=0.5, \tau = 0.02$), defined in Equations~\ref{eq:physical_loss} and~\ref{eq:mixed_loss} respectively. These additional results can be found in Section~\ref{sec:scalability_study}.

\paragraph{Unbalance Error} This is a node-wise physical model loss~\cite{donon2020NeuralNetworksPower}, defined by the unbalanced power at each node, i.e. violation of the Kirchhoff's law. This is a self-supervised loss function since it does not require the actual values of every node as presented in Equations~\ref{eq:physical_loss1}-\ref{eq:physical_loss}.
\begin{equation}
    \Physical{(\hat{X})} = \frac{1}{|\mathcal{N}|}\sum_{i\in\mathcal{N}}{||\Delta P_i}||^2_2 + ||\Delta Q_i||^2_2
    \label{eq:physical_loss1}
\end{equation}
\begin{equation}
    \Delta P_i = \hat{P}_i - \Re\left[\sum_{j\in\mathcal{N}(i)}{
            \Dot{V}_i\left(\frac{\Dot{V}_j-\Dot{V}_i}{z_{i,j}}\right)^*
        }\right] 
        \label{eq:physical_loss2}
\end{equation}
\begin{equation}
    \Delta Q_i = \hat{Q}_i - \Im\left[\sum_{j\in\mathcal{N}(i)}{
            \Dot{V}_i\left(\frac{\Dot{V}_j-\Dot{V}_i}{z_{i,j}}\right)^*
        }\right]
    \label{eq:physical_loss}
\end{equation}
The unbalance error model loss function is purely based on the PF equations where $\mathcal{N}(i)$ is the set of neighbor nodes of node $i$, $\Dot{V}_i=\hat{V}_i\angle \hat{\theta}_i$, $(\hat{V}_i,  \hat{\theta}_i, \hat{P}_i, \hat{Q}_i)$ is the $i$-th row of $\hat{X}$, and $r_{i,j}$ and $x_{i,j}$ refer to the resistance and reactance of line $(i,j)$, respectively.

\paragraph{Mixed Loss}
Finally, we propose the Mixed model loss (Equation~\ref{eq:mixed_loss}) that combines the unbalance error (physical) and MSE loss to efficiently incorporate the PF equations into PowerFlowNet. 
In detail, the Mixed loss for the whole graph with predicted feature matrix $\hat{X}$ and actual real feature value matrix $Y$ is defined as:
\begin{equation}
   \Mixed{(Y, \hat{X})} = 
   w \cdot \MSE{(Y, \hat{X})} + \tau \cdot (1 - w)\cdot \Physical(\hat{X})
   \label{eq:mixed_loss} 
\end{equation}
where $w \in [0,1]$ is a weight to balance the MSE and physical loss, and $\tau$ is a scaling factor bringing the physical loss to the same order of magnitude as the MSE loss. In practice, $w$ and $\tau$ are hyperparameters that can be tuned during training. 

\paragraph{Evaluation}
During the evaluation, we compare the Masked L2 loss as a metric\footnote{We also experimented with training with Masked L2 loss, which resulted in almost the same performance as training with MSE loss.}. The execution times of each implemented method were examined using high-performance hardware configurations, including the ``AMD RYZEN 7 5700X 8-Core'' processor, ``NVIDIA RTX 3060 TI 8GB'' graphics card, and 32GB RAM, ensuring accurate and efficient measurements. Notably, it is important to emphasize the reproducibility of the experiments, as the availability of datasets, the PowerFlowNet framework, and the trained models provided ensure the feasibility of replicating the study's outcomes and procedures. 

\subsection{Performance Comparison}

In this comparison study, PowerFlowNet is evaluated against other baseline methods for PF approximation across three distinct cases, as outlined in Table~\ref{tab:perf-results}. The performance assessment of each approach is conducted based on accuracy, quantified by employing the Masked L2 Loss, as well as the execution time on the same computing machine. The datasets used for evaluation were generated using the Newton-Raphson method, recognized for its accuracy in PF analysis, which we assume to have negligible error for these particular cases. Note that we did not enforce the generator reactor power limits in the Newton-Raphson solution. As expected, the Newton-Raphson method has increased execution time as the scale of the PF problem grows. The second baseline method considered is the DCPF approach, which relaxes the nonlinear problem to a linear one, thereby serving as an upper-performance boundary for the comparison. However, the DCPF only calculates each bus' voltage angle and active power since it assumes the voltage magnitude to be constant across the grid.
\begin{table}[!t]
\centering
\small
\caption{Performance comparison of PowerFlowNet.}
\resizebox{0.485\textwidth}{!}{%
\begin{tabular}{@{}c cc cc cc@{}}
\toprule
Case & \multicolumn{2}{c}{14} & \multicolumn{2}{c}{118} & \multicolumn{2}{c}{6470rte} \\ \midrule
Algorithms & \begin{tabular}[c]{@{}c@{}}Masked \\  L2 Loss\end{tabular} & \begin{tabular}[c]{@{}c@{}}Time\\ (ms)\end{tabular} & \begin{tabular}[c]{@{}c@{}}Masked \\  L2 Loss\end{tabular} & \begin{tabular}[c]{@{}c@{}}Time\\ (ms)\end{tabular} & \begin{tabular}[c]{@{}c@{}}Masked \\  L2 Loss\end{tabular} & \begin{tabular}[c]{@{}c@{}}Time\\(ms)\end{tabular} \\ \midrule
Newton-Raphson & $\approx 0$ & $17.0$ & $\approx 0$ & $20.0$ & $\approx 0$ & $580.0$ \\
DC Power Flow & $45.74$ & $8.0$ & $ 99.87$ & $10.0$ & $510.5$ & $30.0$ \\ 
Tikhonov Reg.  & $2.838$ & $0.4$ & $1.916$ & $0.4$ & $2.934$ & $6100$\\
3-Layer MLP & $0.034$ & $0.6$ & $0.462$ & $0.6$ & $0.590$ & $0.6$ \\
3-Layer GCN & $0.239$ & $1.0$ & $1.257$ & $1.0$ & $1.957$ & $2.0$ \\
PowerFlowNet & $\textbf{0.002}$ & $4.0$ & $\textbf{0.022}$ & $4.0$ & $\textbf{0.303}$ & $4.0$ \\ \bottomrule
\end{tabular}%
}
\label{tab:perf-results}
\end{table}
PowerFlowNet exhibits significantly lower values for the Masked L2 loss when compared with the DCPF method. Additionally, our proposed approach demonstrates significantly reduced execution time~\footnote{Note that when evaluating NN approaches, the training time is not considered since the model needs to be trained only once.} compared to the Newton-Raphson method, with speed improvements of $4\times$ in the 14-node case, $5\times$ in the 118-node case, and $145\times$ in the 6470rte case. Notably, owing to the GNN architecture of PowerFlowNet, the execution time remains consistently low across different cases. This inherent scalability characteristic of GNNs represents a significant advantage over traditional methods, enabling the real-time approximation of PF in extensive power networks. 
Note that the computation time of Newton-Raphson is acceptable ($\approx0.6$s) when computing a small number of cases but might not be enough when thousands of computations are needed in tasks such as planning where the uncertainty of distributed energy resources is introduced. For example, the Newton-Raphson method would take around $20$ minutes to calculate $2000$ PF cases for the 6470rte grid, while PowerFlowNet would only need around $1.3$ minutes.

Furthermore, a comprehensive evaluation of alternative traditional ML approaches was conducted to ascertain the unique benefits of utilizing PowerFlowNet. Initially, the Tikhonov regularizer~\cite{bhlitem137258}, a method designed to enforce smoothness while considering the underlying graph structure, was assessed. However, as indicated in Table~\ref{tab:perf-results}, its performance consistently fell behind the other ML methods in terms of loss across all cases, as well as in execution time within the context of large-scale power networks. Subsequently, a three-layer MLP and a simplified GNN consisting of three layers of GCNs~\cite{DBLP:conf/iclr/KipfW17} were evaluated. Although neither approach surpassed PowerFlowNet in terms of loss (MLP was 16$\times$ worse and the GCN 12$\times$ worse in the 14-case), it is noteworthy that the MLP exhibited a substantially faster execution time, differing by an order of magnitude ($\approx 6.5 \times$ faster).

\subsection{Denormalized Error Distributions}
\begin{figure*}[!ht]
    \centering
    \includegraphics[width=\linewidth]{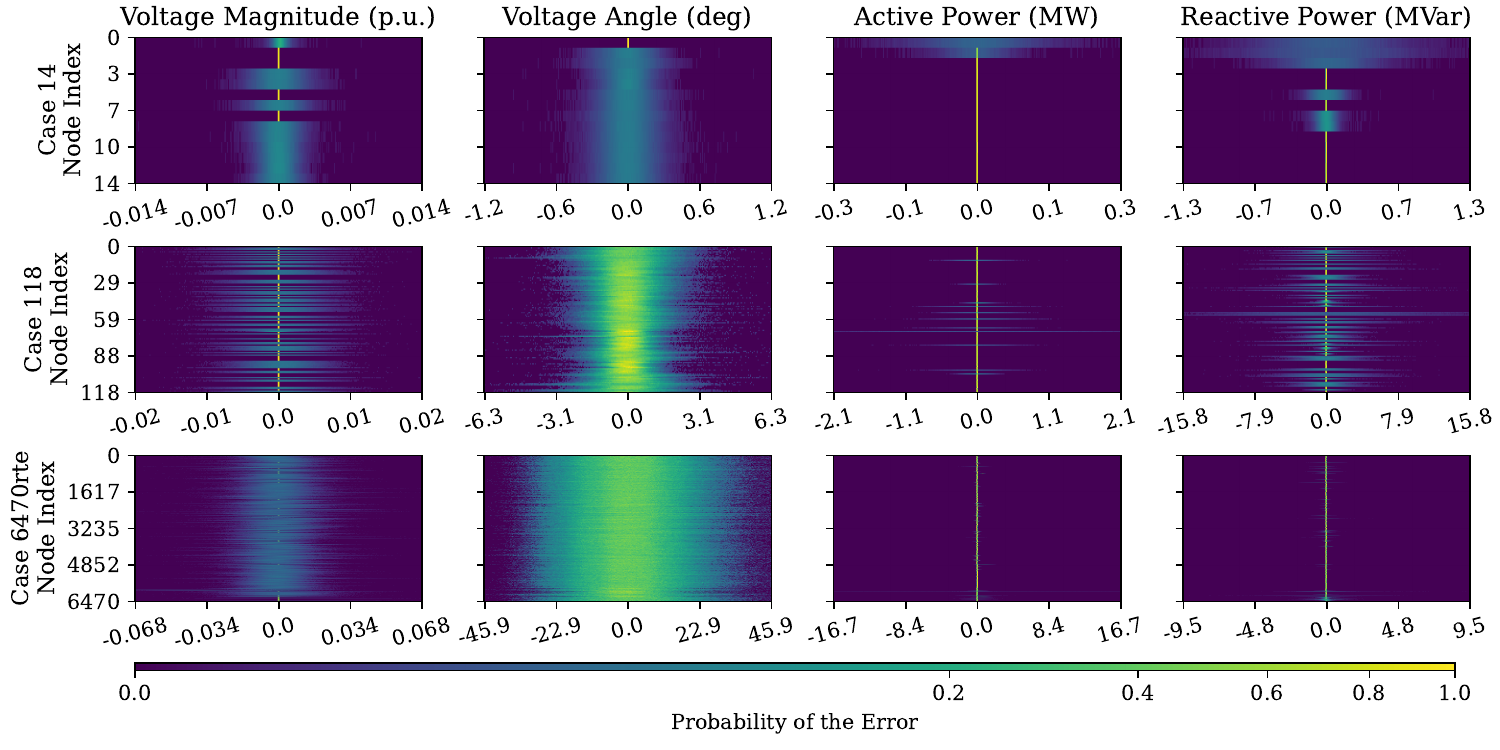}
    \vspace{-3mm}
  \caption{Probability density function of the actual not-normalized error for every node in the test dataset. The brighter the color, the highest the probability of the prediction error of PowerFlowNet being in that region.}
  \label{fig:error_table}
\end{figure*}
After showcasing the performance of PowerFlowNet in three different power networks, it is important to delve deeper into the results and visualize the denormalized error distributions. 
This analytical approach allows us to gain valuable insights into the areas where more accurate predictions can be achieved, as well as the conditions under which these improvements are most evident. Such insights hold practical significance for power systems operators, enabling them to leverage PowerFlowNet effectively in real-world scenarios involving network operation and strategic planning.
Table~\ref{tab:Error-distributions} shows the average error distribution for every one of the four predicted values\footnote{The error distributions include only the predictions of the unknown values.}.

\begin{table}[!t]
\centering
\caption{Average Absolute Denormalized Error}
\label{tab:Error-distributions}
\resizebox{0.485\textwidth}{!}{%
\begin{tabular}{@{}clcll@{}}
\toprule
 \multicolumn{1}{c}{Case}& \multicolumn{1}{c}{$V^m$ (p.u.)} & $\theta$ (\textit{deg})& \multicolumn{1}{c}{$P$ (MW)} & \multicolumn{1}{c}{$Q$ (Mvar)} \\ \midrule
14      & $0.0008 \pm 0.0006$ & $0.09 \pm 0.07$ & $0.02 \pm 0.03$  & $0.13 \pm 0.16$ \\
118     & $0.0022 \pm 0.0018$ & $0.64 \pm 0.54$ & $0.18 \pm 0.36$  & $1.20 \pm 1.47$ \\
6470rte & $0.0072 \pm 0.0061$ & $10.2 \pm 7.89$ & $0.23 \pm 0.42$  & $0.12 \pm 6.46$ \\ \bottomrule
\end{tabular}%
}
\end{table}

Notice that the proposed model achieves precise predictions of p.u. voltage magnitude for all cases, exhibiting an average error of $8\times10^{-4}$ p.u. in the 14-node scenario and $72\times10^{-4}$ in the notably more intricate 6470rte case, all while maintaining a consistently low standard deviation across each case. Additionally, the average absolute error remains minimal for both active and reactive power with respect to the network's actual demand and load, with an actual error of $20$ kW for active power and $130$ kvar for reactive power in the best-case scenario. Nevertheless, it is important to note a significant rise in the standard deviation of predicted reactive power error, particularly in the 6470rte case, where it reaches as high as $6460$ kvar. Similarly, the voltage angle prediction error is barely noticeable in the 14-node case, being only $0.09$ degrees, but increases significantly to $10.2$ degrees in the 6470rte case. This occurs because the range of the voltage angles across larger networks network, i.e., the 6470rte case, is much greater.

Furthermore, we can observe the per-node denormalized error distribution for every case in Fig.~\ref{fig:error_table}. Notice that the prediction error for nodes where we already know the value is practically zero (indicated by the vertical yellow line). Although the average voltage angle error is higher for the 6470rte case ($10.2$ degrees as shown in Table~\ref{tab:Error-distributions}), the average voltage magnitude error is in the same range (less than $0.008$ p.u.) as in the smaller networks, ensuring that the developed PowerFlowNet is accurate enough for fast checking of operational constraints for large numbers of scenarios regardless of network size.

\subsection{Sensitivity to the Hop Size}

To gain insights into the locality of information used by PowerFlowNet to make predictions, we aimed to investigate the potential of using subgraphs as a more scalable and generalizable approach to tackle the challenges posed by the complex PF prediction task.
\begin{figure}[t]
    \centering
    \includegraphics[width=\linewidth]{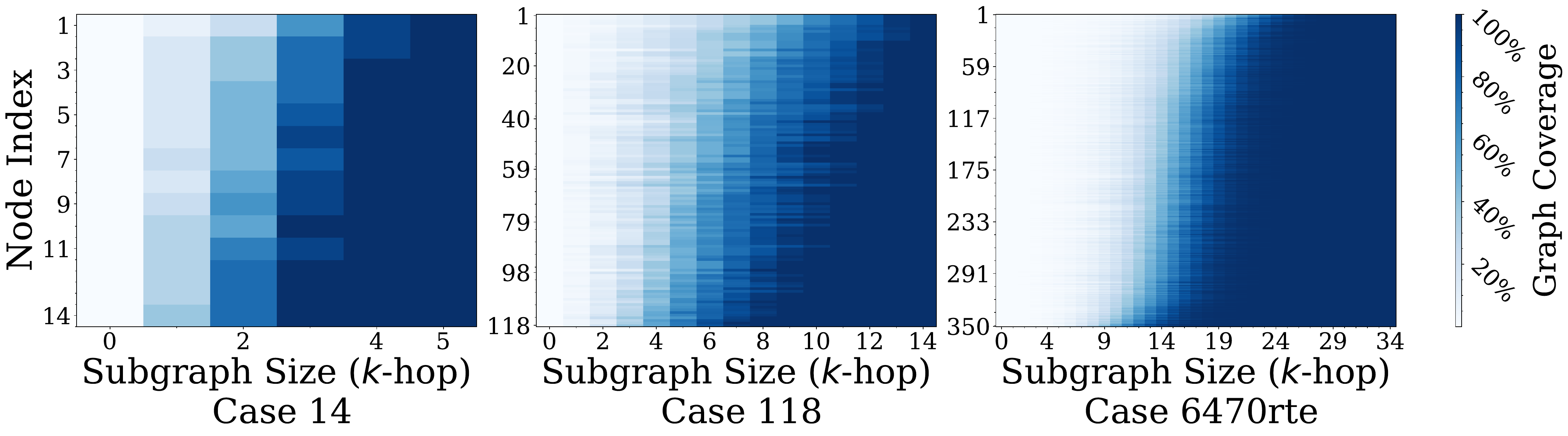}
    \vspace{-3mm}
    \caption{Illustration of graph coverage defined as the number of nodes included in a subgraph, plotted per K-hop size and sorted from slowest to fastest growth for every node.}
    \label{fig:subgraph_growth}
\end{figure}

In this study, we define the receptive field as
the longest distance between any two nodes that can exchange information. Naturally, a larger $K$ and $L$ would increase the receptive field. 
We examined $K$-hop subgraphs across all three cases, where, for reference, a PowerFlowNet model with $K=3$ and four layers of TAGConv ($L=4$), a straightforward computation yields an effective receptive field spanning 12 nodes. In practice, the size of these receptive fields, measured in the number of nodes, will vary based on the specific graph topology.
To better understand this topological effect, we looked at the speed at which K-hop subgraphs grow from each node in the graph, also defined as graph coverage. In detail, for each node, we made $K$-hop subgraphs for K between 1 and the graph diameter and measured the subgraph size after each hop, as shown in Fig.~\ref{fig:subgraph_growth}. The graph diameter is the longest distance between any pair of nodes within a graph, representing the maximum distance between any two points in the network.
This indicates the distribution of receptive field sizes seen during training; interestingly, it appears to form a Gaussian shape, particularly visible in the largest graph in the dataset with $6470$ nodes.
As expected, the larger the graph size the more hops are required to cover the whole graph. For example, $4$ hops are required in the 14-node over $6$ hops in the 118-node case, and $11$ to $25$ hops in the 6470rte case.

\begin{figure}
    \centering
    \includegraphics[width=\linewidth]{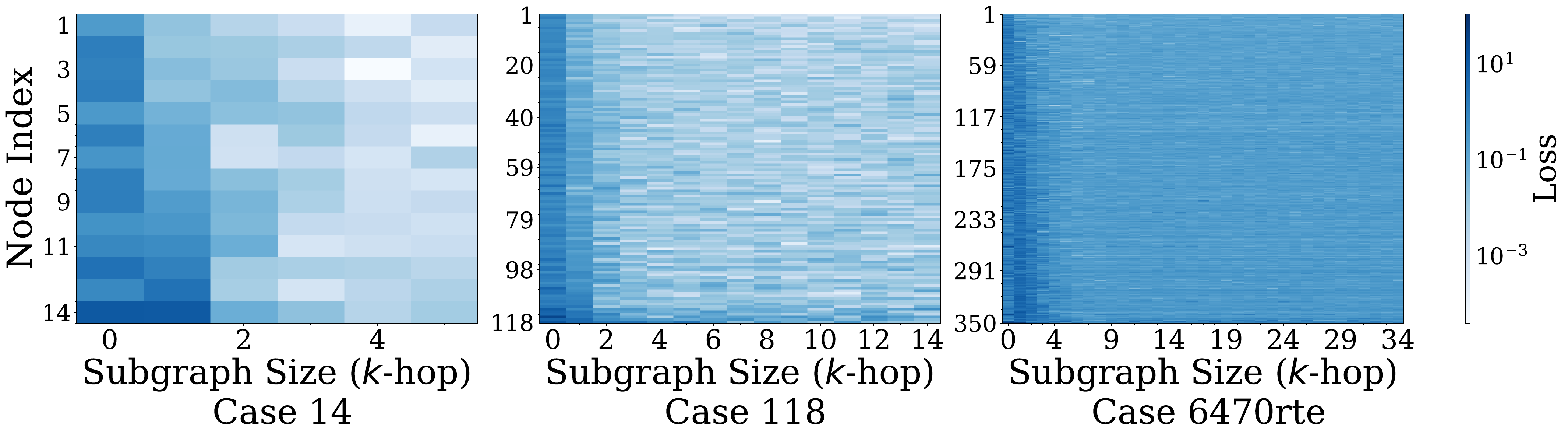}
    \vspace{-3mm}
    \caption{Masked L2 loss on the central node of a $k$-hop subgraph for each node and $k$, sorted by the total node loss.}
    \label{fig:loss_per_hop_per_node}
\end{figure}

Afterward, those subgraphs were each passed through the PowerFlowNet model, and the Masked L2 loss was calculated only on the node from which the $K$-hop neighborhood was generated.
This showed us the actual loss distribution among nodes depending on how far a $K$-hop neighborhood can reach. The detailed results are illustrated in Fig.~\ref{fig:loss_per_hop_per_node}.
We noticed that this loss distribution has a different shape from the subgraph growth figure, and indeed there is variation in which nodes are more or less affected by a decrease in subgraph size. 

Interestingly, we can observe that only around $3$ hops are required to obtain the minimum loss for most of the nodes, experimentally proving the reason TAGConv of order $K=3$ can achieve high-quality results. This also suggests that for most buses (nodes) in a power network, we could make PowerFlowNet predict the PF values even faster by only looking at smaller connected parts.

\begin{figure}
    \centering
    \includegraphics[width=\linewidth]{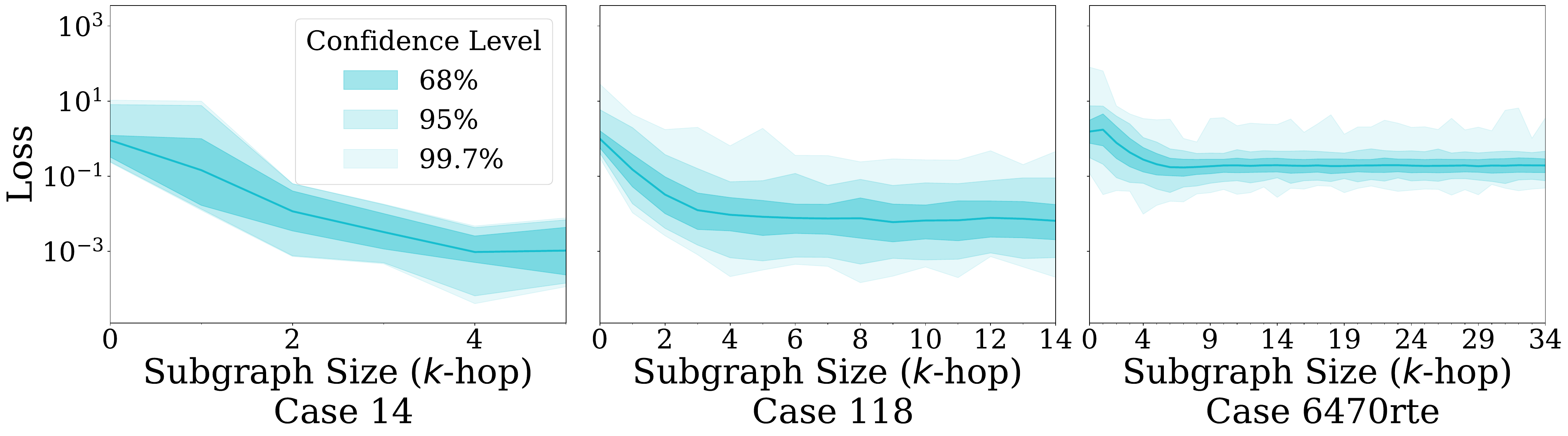}
    \vspace{-3mm}
    \caption{Average Masked L2 loss on the central nodes of a $k$-hop subgraph for different numbers of $k$.}
    \label{fig:loss_per_hop}
\end{figure}

Finally, we plot the average loss across nodes in Fig.~\ref{fig:loss_per_hop} and discover that neighborhood hop counts well below the effective receptive field's hop count are enough to make predictions practically identical to those when using the full graph.
This indicates that local predictions in a large network are feasible.
Most importantly, we argue that training on larger graphs could be made more efficiently by sampling subgraphs across different graphs available in the dataset, so as to make sure that stochastic gradient descent steps are performed using batches that capture more of the variation available in the dataset. Subsequently, this approach has the potential to augment the scalability aspects of PowerFlowNet.


\subsection{Network Generalization Capabilities}
\label{sec:scalability_study}
We demonstrate that our model has valuable generalization features to perform well in different network topologies after training in one particular topology. Also, we investigate how PowerFlowNet's performance scales with the NN's model size, the size of the training dataset, and the model loss function used. 

Based on the standard setting PowerFlowNet (Medium) with $L=4, h=128$, we created two variants by modifying the total number of the model's trainable parameters by one order of magnitude as shown in Table~\ref{tab:scalability-performance}. In detail, we created PowerFlowNet (Small) with $L=2, h=64$ and PowerFlowNet (Large) with $L=5, h=512$. Table~\ref{tab:scalability-performance} shows the relation between the model size and the Masked L2 loss when trained with different model loss functions for the 118-node case. We note that the performance scales very well with the model size. Most importantly, from the medium to large model, the Masked L2 loss is almost one order of magnitude lower, signifying the learning potential of PowerFlowNet. 

\begin{table}[!t]
    \centering
    \small
    \caption{Masked L2 loss of scaled models on the 118 case.}
    \resizebox{0.485\textwidth}{!}{%
    \begin{tabular}{ccccc}
        \toprule
       \birow{Model} & \birow{\# Params}& &Trained with &   \\
        \cmidrule{3-5}
        && MSE & Physical & Mixed \\
        \midrule
        \mc{1}{l}{PowerFlowNet (Small)} & $32$k    & $0.079$  & $0.757$  & $0.109$ \\
        \mc{1}{l}{PowerFlowNet}& $357$k   & $0.022$  & $0.667$  & $0.057$ \\
        \mc{1}{l}{PowerFlowNet (Large)} & $7375$k  & \textbf{0.002}  & \textbf{0.628}  & \textbf{0.004} \\
        \bottomrule
    \end{tabular}
    }
    \label{tab:scalability-performance}
\end{table}

The model performance does not differ greatly when trained with physical loss (as used in~\cite{donon2020NeuralNetworksPower}). This means learning with physical loss is challenging regardless of the model's capability. However, we notice that the Masked L2 performance difference between training with MSE and mixed loss shrinks as we move to the large model. This suggests that this capability gain allows us to optimize both the MSE and the physical loss, thus making precise and physics-conforming predictions. 

\begin{table}[!t]
    \centering
    \small
    \caption{Performance evaluation for varying training cases. 
    }
    \resizebox{0.485\textwidth}{!}{%
    \begin{tabular}{cccccccc}
        \toprule
        \multicolumn{1}{l}{Eval. Case}  && \multicolumn{3}{c}{118} & \multicolumn{3}{c}{14} \\
        \midrule
        \multicolumn{1}{l}{Eval. Metric} && M. L2 & MSE & Phys. & M. L2 & MSE & Phys. \\
        \midrule
                                            & \mc{1}{l}{118+14 (L)}  & \textbf{0.0023}  & \textbf{0.0011}  & \textbf{4.9676}  & \textbf{0.0014}  & \textbf{0.0006}  & \textbf{0.1336} \\
        \multicolumn{1}{l}{Train. Case}   & \mc{1}{l}{118 (L)}     & $0.0029$  & $0.0014$  & $5.4080$ & $6.1998$  & $2.5947$  & $40.942$ \\
                                            & \mc{1}{l}{118 (M)}     & $0.0220$  & $0.0105$  & $58.9221$ & $3.7860$  & $3.7860$  & $45.426$ \\
        \bottomrule
    \end{tabular}
    }
    \label{tab:large-model-cap}
\end{table}

To further test PowerFlowNet's capability, we trained the large model on a mixed dataset composed of the 14 and the 118 cases (118 + 14). We unified the training loss function with MSE loss and evaluated the trained models using different metrics (MSE, Masked L2, and physical loss) on each of the two cases individually. The results are given in Table~\ref{tab:large-model-cap}. After comparing with training large (L) and medium (M) models on the 118-node case, we noticed that even without explicitly bringing physical model loss in training, the large models achieve better performance. Large models are around $10\times$ better in terms of unbalance error when trained and evaluated on the 118-node case. When we augment the 118-node case training set with the 14-node case, the Large PowerFlowNet model performs better than when trained only in the 118-node case ($0.0006$ less Masked L2 loss), achieving an excelling accuracy in the 14-node case ($0.0006$ less Masked L2 loss than just training in the 14 case). This suggests that the proposed large PowerFlowNet model can operate on multiple graphs and learn the underlying physics without sacrificing accuracy. 

\subsection{Architectural Ablation Study}

Finally, architectural ablation experiments were carried out to demonstrate the vitality of PowerFlowNet's unique structures through component removal. Therefore, the experiments were designed to highlight how each PowerFlowNet component contributes to generating high-quality predictions.
Table~\ref{tab:modules} depicts the results of these experiments. Table~\ref{tab:modules} is split into four variations of PowerFlowNet: the full model, PowerFlowNet with $L=1$, PowerFlowNet without message passing, and PowerFlowNet with $L=1$ and without message passing. Then, we can observe how the model performs in terms of Masked L2 loss in all different scenarios.

\begin{table}[!t]
\caption{PowerFlowNet component significance analysis}
\label{tab:modules}
\centering
\begin{tabular}{@{}ccccc@{}}
\toprule
Case & \begin{tabular}[c]{@{}c@{}}Full \\ Model\end{tabular} & \begin{tabular}[c]{@{}c@{}}Model \\ 1-Layer\end{tabular} & \begin{tabular}[c]{@{}c@{}}No Message\\  Passing (MP)\end{tabular} & \begin{tabular}[c]{@{}c@{}}1-Layer \& \\ No MP\end{tabular} \\ \midrule
14 & $\textbf{0.002}$ & $0.11$ & $0.78$ & $0.08$ \\
118 & $\textbf{0.022}$ & $0.38$ & $1.01$ & $0.24$ \\
6470rte & $\textbf{0.303}$ & $1.12$ & $1.91$ & $0.88$ \\ \bottomrule
\end{tabular}%
\end{table}

 

As observed, none of the PowerFlowNet's variations performed as well as the complete model. In the 1-layer case, it is visible that the model fails to capture distant node dependencies, resulting in up to $55\times$ higher error. Similarly, the model performs poorly (up to $390\times$ higher error) in the no message passing case since the edge features are completely ignored, highlighting the importance of incorporating the edge features. On the other hand, the 1-layer model without message passing performs the best (up to $40\times$) since its functionality is similar to stacked MLPs. Consequently, this analysis shows that every part of PowerFlowNet is important in achieving high-quality PF approximations.

\section{Conclusions} \label{chap6}

In this paper, we presented PowerFlowNet, a novel data-driven algorithm that capitalized on the efficiency and capabilities of GNN operations applied to the power network's topology, leading to a accurate approximation of the PF. Specifically, our approach transformed the traditional PF problem into a GNN node-regression task by representing buses as nodes and transmission lines as edges while preserving network connectivity. 
Our model featured a distinctive configuration involving a mask encoder in conjunction with a sequence of our proposed PowerFlowConv layers, designed to aggregate features across the entire graph and inherently learn the dynamics of the underlying PF.
The results verified PowerFlowNet's ability to generate high-quality solutions for a diverse set of network cases including the 6470rte very large-scale power network of France. Most importantly, our method produces high-quality PF predictions at a fraction of the time of traditional solvers, regardless of the scale of the network, outperforming the well-established Newton-Raphson method in terms of execution time while having closely comparable results.
Ultimately, our ablation studies reveal the robust aspects and potential limitations of our approach, thereby enabling the versatile deployment of PowerFlowNet across a wide array of real-world power system operation and planning scenarios.




\bibliographystyle{IEEEtran}

\bibliography{IEEEabrv,ref.bib,references.bib}


\end{document}